\documentclass[11pt,a4paper]{article}
\usepackage{times,latexsym}
\usepackage{url}
\usepackage[T1]{fontenc}
\usepackage{enumitem}
\usepackage{amsmath,amsfonts,amssymb}
\usepackage{graphicx}
\usepackage{booktabs}
\usepackage{natbib}
\usepackage[colorlinks=true,linkcolor=blue,citecolor=blue,urlcolor=blue]{hyperref}

\usepackage[margin=1in]{geometry}

\title{The Medical Metaphors Corpus (MCC)} 

\author{
  Anna Sofia Lippolis \\
  University of Bologna \\
  Bologna, Italy \\
  \texttt{annasofia.lippolis2@unibo.it}
  \and
  Andrea Giovanni Nuzzolese \\
  CNR Institute for Cognitive Sciences and Technologies \\
  Bologna, Italy \\
  \texttt{andrea.nuzzolese@istc.cnr.it}
  \and
  Aldo Gangemi \\
  University of Bologna \\
  Bologna, Italy \\
  \texttt{aldo.gangemi@unibo.it}
}

\date{}

\begin{document}
\maketitle
\begin{abstract}
Metaphor is a fundamental cognitive mechanism that shapes scientific understanding, enabling the communication of complex concepts while potentially constraining paradigmatic thinking. Despite the prevalence of figurative language in scientific discourse, existing metaphor detection resources primarily focus on general-domain text, leaving a critical gap for domain-specific applications. In this paper, we present the Medical Metaphors Corpus (MCC), a comprehensive dataset of 792 annotated scientific conceptual metaphors spanning medical and biological domains. MCC aggregates metaphorical expressions from diverse sources including peer-reviewed literature, news media, social media discourse, and crowdsourced contributions, providing both binary and graded metaphoricity judgments validated through human annotation. Each instance includes source-target conceptual mappings and perceived metaphoricity scores on a 0-7 scale, establishing the first annotated resource for computational scientific metaphor research. Our evaluation demonstrates that state-of-the-art language models achieve modest performance on scientific metaphor detection, revealing substantial room for improvement in domain-specific figurative language understanding. MCC enables multiple research applications including metaphor detection benchmarking, quality-aware generation systems, and patient-centered communication tools. \end{abstract}

%\iftaclpubformat

\section{Introduction}

Metaphor is a fundamental cognitive mechanism that structures how humans categorise experience and reason about abstract domains.  Everyday communication is saturated with metaphoric expressions: it suffices to think about when we describe a \emph{heated} debate or conceptualize time as a \emph{resource}.  Lakoff and Johnson’s \emph{Conceptual Metaphor Theory} formalised this insight, arguing that linguistic metaphors reflect systematic mappings between a \emph{source} domain and an \emph{target} domain \citep{lakoff1981metaphors}. Four decades of psycholinguistic evidence have confirmed that such mappings influence thought and behaviour \citep{thibodeau2019metaphor,robins2000metaphor}.  For instance, framing climate change as a \emph{war} elicits greater urgency and pro‐mitigation intent than framing it as a \emph{race} \citep{flusberg2017metaphors}, while the choice between \emph{fighting a battle} and \emph{navigating a maze} in oncology discourse measurably affects patients’ emotional response and treatment decisions \citep{semino2018integrated}.

Metaphor is pervasive even in the most technical-words-filled genres: corpus studies estimate that \textasciitilde{}11--15\% of propositions in peer‐reviewed research articles involve figurative language \cite{cameron2003metaphor,low2008metaphor}.  Yet precisely these high‐stakes domains expose severe blind spots in current language technologies. Despite advances in large language models (LLMs), figurative language understanding remains brittle \citep{stowe2021metaphor,leivada2023sentence}.  Recent evaluations show that LLMs excel at proportional analogies \citep{webb2023emergent} but struggle with higher‐order relations such as metaphor, especially when associative cues must be suppressed \citep{wijesiriwardene2023analogical,stevenson2023large}. The gap is unsurprising: most models are trained on surface‐level co‐occurrence statistics rather than cognitively grounded representations \cite{schrimpf2021neural,rule2020child}. These cues tend to be most evident in domain-specific metaphors rather than generic ones, thus medical metaphors can serve as additional test cases for such scenarios.%why + difficult to find literality

A major impediment to using varied domain-derived metaphors for computational experiments is data scarcity.  Existing benchmarks either target isolated lexical metaphors or everyday conceptual metaphors rooted in news and fiction (see Section 2). To our knowledge, no publicly available resource offers fine‐grained annotations of domain‐specific metaphors in scientific writing.  Likewise, downstream applications such as clinical decision‐support and patient‐centric text generation lack training data that distinguishes conventional metaphors from perceivedly novel ones.

Furthermore, studies show metaphoricity is a range rather than a binary label, however, this continuum has not been usually annotated in metaphor datasets \citep{julich2023zooming,bisang2006corpus,dunn2010gradient,gibbs2015counting}. Existing evaluation frameworks treat all human annotations equally, despite varying levels of annotator consensus. A model that fails to detect metaphors with high human agreement represents a more serious limitation than one that struggles only with cases where human annotators themselves show substantial disagreement.

To bridge this gap, we introduce the \textbf{Medical Metaphors Corpus} (MCC), a 792‐item dataset that aggregates medical, health and disease metaphors from nine sources across heterogeneous channels from scholarly articles to social media, and enriches each sentence with crowd‐validated ratings of perceived metaphoricity.  

By providing the first discourse‐aware, domain‐balanced resource of this kind, we enable systematic testing of LLMs’ domain-specific metaphor competence, support contrastive studies between expert and lay framing, and lay empirical foundations for applications ranging from claim mining to the generation of patient‐friendly explanations. In this context, we also propose the use of confidence-weighted evaluation metrics that prioritize items with stronger human consensus while de-emphasizing controversial cases.

The remainder of this paper is organised as follows: Section 2 reviews existing metaphor datasets and computational approaches. Section 3 details our data collection methodology. Section 4 presents our annotation framework and quality control measures. Section 5 provides comprehensive dataset statistics and disagreement analysis. Section 6 evaluates state-of-the-art language models on metaphor detection. Section 7 discusses implications for computational metaphor processing and scientific communication tools.

\section{Background}\label{sec:background}
This section describes the background to our approach to curating domain-specific metaphor instances from peer-reviewed literature, establishing the theoretical foundation for scientific metaphor annotation and computational use.

\subsection{Conceptual Metaphor Theory and Scientific Rhetoric}
Lakoff and Johnson’s Conceptual Metaphor Theory (CMT) foregrounds metaphor as a cognitive mechanism composed of a source and a target domain that structures abstract reasoning \citep{lakoff1981metaphors}. Recent work in the philosophy of science shows that tracking metaphor evolution through the lens of CMT offers insight into how entire research programmes shift over time, revealing hidden argumentative moves and disciplinary cross-fertilisation \citep{szymanski2019remaking}. For instance, corpus studies of COVID-19 discourse demonstrate how \textsc{WAR}, \textsc{JOURNEY}, and \textsc{NATURAL DISASTER} frames circulate to legitimise policy and sway public sentiment \citep{alkhammash2023bibliometric}.  Pedagogical research argues that explicit metaphor analysis fosters scientific literacy and civic responsibility in students \citep{taylor2018problem}.  Outside biomedicine, financial linguistics exposes how shared metaphors (e.g.\ \textsc{risk is enemy}) constrain regulatory thinking \citep{young2001risk}.

\subsection{Metaphor datasets}
The Master Metaphor List \cite{lakoff1991master} marked a crucial milestone by compiling over 791 conceptual metaphor mappings, creating the first comprehensive evaluation benchmark. \citet{mason2004cormet}'s CorMet system represented the first large-scale corpus-based approach to metaphor extraction, dynamically mining Internet corpora using selectional preference patterns. The development of reliable annotation schemes proved crucial for creating high-quality metaphor datasets. The \citet{steen2002towards}'s MIP (Metaphor Identification Procedure) \citep{steen2019chapter} provided the first explicit, systematic method for identifying metaphorical word usage. MIPVU (Metaphor Identification Procedure VU University), refined and extended MIP with more detailed guidelines for borderline cases. The VU Amsterdam Metaphor Corpus \citep{steen2010vu} became the field's primary benchmark, containing approximately 190,000 lexical units from the BNC-Baby subset. The LCC Metaphor Datasets \citep{mohler2016introducing} represented a leap in scale and linguistic diversity. 
MetaNet is a multilingual metaphor repository and computational system that systematically identifies and analyzes generic conceptual metaphors, partly derived by the Master Metaphor List, across domains using formalized frames and semantic mappings. The project builds on CMT to create structured networks of searchable metaphors spanning English, Spanish, Persian, and Russian \citep{metanet} . \citet{gangemi2018amnestic} extend MetaNet's framework with the Amnestic Forgery ontology, which reuses and enhances the MetaNet schema through integration with Framester to address both semiotic and referential aspects of metaphorical mappings. Amnestic Forgery demonstrates how MetaNet's structured approach can support automated metaphor generation and ontological reasoning about figurative language.
Recent developments have emphasized multimodal and multilingual expansion. The MultiCMET dataset \citep{zhang2023multicmet} provides 13,820 text-image pairs from Chinese advertisements, representing the first large-scale multimodal metaphor dataset in Chinese. 
The MUNCH (Metaphor Understanding Challenge Dataset) \citep{tong2024metaphor} provides over 10,000 paraphrases plus 1,500 inapt paraphrases, representing the first comprehensive benchmark for evaluating large language model metaphor understanding.
Multimodal metaphor processing has emerged as a crucial frontier. The MET-Meme dataset \citep{xu2022met} enables cross-modal metaphor analysis.

\subsubsection{Domain-Specific Resources for Medical Metaphor}
Figurative language in specialised medical prose is under-resourced. \citet{semino2018integrated} annotated more than one million cancer-forum posts for metaphor use and patient affect, but the dataset is not currently available for people not registered at an institution outside the UK. The \#ReframeCovid initiative crowdsourced pandemic metaphors but lacked sentence-level gold labels \citep{olza2021reframecovid}.  Inventories such as \citet{van1997metaphors}’s conceptual medical metaphors appendix and \citet{metamia}’s crowd-sourced metaphors  offer numerous raw annotated examples yet remain heterogeneous. The MCC dataset aims to unify these strands into a unique annotated daraset.

\subsection{Computational Metaphor Detection and Interpretation}
Early neural models targeted lexical metaphor; MelBERT’s late-interaction architecture remains a strong baseline on MOH-X, VUA and TroFi datasets \citep{choi2021melbert}.  Frame-informed detectors such as FrameBERT \citep{li2023framebert} improve interpretability by aligning predictions with semantic roles.  At the conceptual level, MetaPRO retrieves and ranks candidate source–target mappings without explicit prompts \citep{mao2023metapro}, whereas theory-guided prompting (\textsc{TSI-CMT}) injects CMT constraints into chain-of-thought reasoning for LLMs \citep{tian2024theory}.  Logic-augmented approaches further enhance multimodal analogical reasoning by binding LLM output to symbolic constraints \citep{gangemi2025logic}, which are being applied, among other tasks, to metaphorical computational processing \citep{de2025neurosymbolic}.

\subsection{Metaphor for Science Communication}
A manually curated ``metaphor menu'' paradigm has been proposed in patient-care settings—offering alternative framings (e.g. \textsc{journey} vs \textsc{battle}) to respect individual preferences and mitigate distress \citep{semino_metaphor_menu}; computational support for curating such menus is still to be implemened.
Another work concerning scientific communication directly targets scientific writing, analyzing metaphor variation in \textit{Nature Immunology} and \textit{New Scientist} articles \cite{semino2018integrated}. Most other resources focus on general, argumentative, or political language. We didn’t find mention of large, domain-specific corpora for scientific metaphors in the included studies.
Computationally, metaphor generation for scientific communication has been recently investigated. Metaphorian pairs GPT-4 with interactive structures to help science writers draft vivid extended metaphors and evaluates candidates for novelty and explanatory power \citep{kim2023metaphorian}.  

These studies confirm metaphor’s rhetorical power and showcase promising detectors, yet they reveal two main gaps: (i) a shortage of harmonised medical datasets and (ii) limited support for controlled metaphor generation.  MCC directly addresses these gaps, furnishing the foundation needed for domain-aware metaphor processing for NLP.

\section{Data Collection}
%The data collection process consisted in merging and harmonizing already available heterogeneous data. The data derived mostly from works that showcased metaphors or examples thereof already available in the literature about metaphors (see Related Works). The metaphors collected in these works come from different sources
%derives from the following sources, mostly from literature that has a sua volta found metaphors:

%Figure~\ref{fig:MCC_pipeline} sketches the three-stage workflow.

Our data collection followed a systematic approach to identify annotated scientific metaphors in existing literature. We conducted searches using keywords: ``scientific metaphor'', ``medical metaphor'', ``biological metaphor'', ``conceptual metaphor AND science'', across major scholarly and linguistic databases (Linguistics and Language Behavior Abstracts, MLA International Bibliography) and computational linguistics venues (ACL Anthology). Sources were included if they: (1) contained explicit sentence-level metaphor annotations in medical or biological domains, (2) provided source-target mappings following CMT framework, and (3) offered sufficient context for metaphoricity assessment. This yielded nine primary sources spanning different discourse types (academic literature, news media, social platforms, patient narratives, crowdsourced data) to ensure genre diversity while maintaining domain focus.

Each source underwent standardization: sentences were extracted verbatim and tagged with provenance information. Pre-existing annotations (source/target domains, metaphor types) were preserved where available to maintain scholarly continuity.

\begin{table}[h]
\centering
\caption{Primary sources for MCC divided by channel (Chan) and number of metaphors (N). For Channels, \textit{Lit} concerns academic literature, \textit{News} the news domain, \textit{SoMe} social media, \textit{Crowd} stands for crowdsourced.}
\label{tab:MCC_sources}
\small
\begin{tabular}{@{}p{4.5cm}cr@{}}
\toprule
\textbf{Source} & \textbf{Chan} & \textbf{N} \\ \midrule
\citet{van1997metaphors} Medical metaphors & Lit & 455 \\
\citet{camus2009variation} UK News & News & 19 \\
\citet{kaikaryte2020conceptual} UK news & News & 145 \\
\citet{semino2018integrated} patient forum & SoMe & 27 \\
\citet{fereralda2022use} cancer stories & SoMe & 35 \\
\citet{cheded2022dead} medical metaphors & News & 35 \\
\citet{gibbs2002embodied} cancer narratives & SoMe & 50 \\
\citet{sinnenberg2018content} diabetes Twitter & SoMe & 40 \\
\citet{metamia} & Crowd & 16 \\ \midrule
\textbf{Total} & & \textbf{792} \\
\bottomrule
\end{tabular}
\end{table}

\subsection{Literature}

In Metaphors in Medical Texts, by \citet{van1997metaphors}, the authors analyze how conceptual metaphors are used in medicine by analyzing scientific articles. In the text, the authors devise 455 conceptual metaphors which are classified into different metaphor categories, source and target domains. 

\subsection{News outlets}
     
\citet{camus2009variation} analyses 19 cancer conceptual metaphors found in The Guardian. \citet{kaikaryte2020conceptual} analyzes conceptual metaphors in popular medical discourse: 145 from popular UK news outlets such as The BBC, The Guardian, or The Daily Mail. The scope of these works is usually to analyze how diseases are talked about in popular discourses from the point of view of CMT.
\citet{cheded2022dead} analyze 35 medical metaphors for understanding the consumption of preventative healthcare in a news setting.

%-Camus, Variation of cancer metaphors in scientific texts and press popularisations

%-CONCEPTUAL METAPHORS IN POPULAR MEDICAL DISCOURSE by Agnė Kaikarytė

\subsection{Social media}

Many works focus on social media discourse of illnesses. In fact, people anonymously can share more freely what they think, and it's a different perspective than one of both ``institutional'' outlets such as news or scientific literature. In this way, it is possible to get a glimpse into what the patient really experiences.

While proposing an integrate approach to metaphor and framing, \citet{semino2017online} selects for presentation 27 metaphors from an UK-based online forum for people with cancer and identifies 35 metaphors apt for discussion about the use of conceptual metaphor in cancer patient stories.
%In Olza et al., the authors collect 630 metaphors of Covid in all languages, sourced from Twitter. We extract a subset of 138 that correspond to both textual and visual metaphors. -- did not do this
\citet{fereralda2022use} present five metaphors in popular discourse online and focuses on the FORCE forum. Finally, \citet{sinnenberg2018content} collect 40 metaphors of diabetes online, on Twitter specifically.

\subsection{Interviews}
\citet{gibbs2002embodied} collect 50 conceptual metaphor from interviews with 6 middle-class women in recovery from cancer.

\subsection{Crowdsourced data}
Metaphors can also be collected from crowdsourced data. In particular, \citet{metamia} is a website where users can freely submit metaphors and analogies found online. They can specify the source and the target of the trope, along with author and link of the source.
The website is not structured by themes but rather has a keyword-based search option. To collect medical metaphors, the following keywords: ``cell'', ``disease'', ``illness'', ``cancer", ``biology'' were searched to filter from inputs by users. Furthermore, these results were manually filtered by an expert according to their actual presence of a metaphor, so the implicit comparison instead of the explicit analogy, and according to the presence of a good example. As a result of this process, we obtain 16 annotated metaphors.

\section{Annotation model}
Given the heterogeneous nature of our source material and the need to capture information beyond basic source-target mappings, we sought to measure the perceived metaphoricity of each expression. This approach addresses two key insights from the literature: metaphoricity exists on a continuum rather than as a binary property, and many medical metaphors are highly conventionalized, potentially affecting their perceived figurativeness.

Thus, we expanded the usual
\textit{source–target} schema with two annotations:

\begin{enumerate}
  \item \textbf{Binary metaphoricity} (\textsc{M}/\textsc{L}).  
  \item \textbf{Perceived metaphoricity scale} (0 = literal … 7 = highly
        metaphorical).
\end{enumerate}

All the metaphors were annotated through a Qualtrics survey upon specific instructions by Twenty-seven advanced students of the \emph{Informatica Umanistica} programme participated (\textasciitilde{}C1 English). To these, 15 online linguists recruited through the Linguistlist newsletter were added, with English as a primary language, making up a total of 42 annotators who annotated about 80 sentences each. Prior to annotation, all participants received instructions on defining metaphor and metaphoricity along with an example. To ensure reliability, each sentence received independent annotations from a minimum of two annotators, with systematic overlap designed to calculate inter-annotator agreement.

For each sentence, two questions were asked: (i) ``Does this sentence contain a metaphor?''; (ii) ``On a scale from 0 (literal) to 7 (very metaphorical), how metaphorical do you perceive this sentence to be?. If you put \textit{No} to the previous question, write 0.''

\section{Quality control and inter-annotator agreement}

The responses were filtered according to consistency of the answer with the yes/no responses: if the metaphor was judged literal, the metaphoricity was explicitly said to be 0. They were also manually checked with respect to the amount of metaphors they were to input. Empty submissions were of course removed.

Fleiss’ kappa was 0.23, with the average percent agreement of 60\%. For the agreement on the Likert scale, average Pearson \textit{r} is 0.4 with the Spearman \textit{p} being 0.4.

\section{Dataset statistics}
Our corpus contains 792 sentences drawn from scientific writing in which metaphorical language is either suspected or confirmed. Within these sentences we identified 82 distinct metaphor types, spanning 24 unique target domains and 38 unique source domains. Each sentence was labelled by at least two annotators, and we derived a gold‐standard label via majority vote together with the mean metaphoricity score for that sentence.

\medskip
\noindent
Across all annotations, ``Yes''/``No'' decisions are distributed as follows 353 ``yes'' (44.57\%), ``305'' no (38.51\%), 134 ties (16.9\%).

A tie occurs when annotators are evenly split; e.g.\ the sentence
\textit{``In theory, blocking any of the necessary steps for invasion listed in Table~7 could prevent tumor cell invasion.''}.

\paragraph{Disagreement Metrics.}
For the binary judgments we quantified disagreement with:

We first compute the proportion of ``yes'' votes, denoted \(p_{\text{yes}}\), and take the remaining fraction \(1-p_{\text{yes}}\) as the ``no'' votes.  
The disagreement score is then defined by
\[
d \;=\; 1 - \bigl|p_{\text{yes}} - (1-p_{\text{yes}})\bigr|
      \;=\; 1 - 2\!\left|p_{\text{yes}} - \tfrac12\right|.
\]
This index ranges from \(0\) when all annotators agree, to \(1\) when the panel splits exactly fifty–fifty.  
Because the formula measures how far the vote share strays from perfect balance and then inverts the scale, larger values indicate stronger discord while smaller values mark stronger consensus.

% (Optional) Unlike $\kappa$‑style statistics, this lightweight index
% does not adjust for chance agreement; it simply reports how polarized
% the vote was on each item.

For metaphoricity‐rating questions (0–7 scale) we used the standard deviation $\sigma$ of the ratings as the disagreement index: higher~$\sigma$ indicates greater annotator divergence about metaphorical intensity.

From the statistical analysis of the dataset, we identified the following key findings:
\begin{itemize}
  \item \textbf{Binary metaphoricity vs.\ Range.} Sentences judged metaphorical receive substantially higher metaphoricity ratings than non-metaphorical ones ($\mu_{\textsc{meta}}=3.41$ vs.\ $\mu_{\textsc{non}}=0.16$, $\Delta=3.25$ points), a pattern that holds for 95\% of question pairs.
  \item \textbf{Boundary cases.} The highest binary disagreement (perfect 50/50 splits) arises in three main situations: 
    (i) scientific terminology with a possible metaphorical reading (e.g.\ \textit{``drug transport''}), 
    (ii) highly lexicalised conventional metaphors, and 
    (iii) domain-specific phrases whose interpretation depends on the context in which they are set.
  \item \textbf{Uncertainty about metaphoricity.} The maximum rating variance observed ($\sigma=4.95$) coincides with these boundary cases, indicating that uncertainty about a sentence’s metaphorical status directly translates into uncertainty about its perceived literality.
\end{itemize}

\medskip
\noindent

\subsection{Metaphoricity}
The rating distribution on metaphoricity shows a heavily skewed pattern toward the lower end of the 0-7 scale. The spike at rating 0 represents roughly 38\% of all ratings and is more than five times larger than any other single rating category.
The distribution suggests a polarized reception, with a substantial group giving the absolute lowest rating while the remaining ratings are more evenly spread across categories 1-7. 

The dataset is anchored by a large corpus from \citet{van1997metaphors}, which exhibits a mean metaphor rating of 2.34. This source likely serves as the backbone of the analysis, offering a representative baseline for the effectiveness of metaphor usage in formal biomedical discourse. In contrast, journalistic sources such as BBC (mean: 2.28), The Guardian (mean: 2.83), and the Telegraph (mean: 1.99) cluster around slightly lower to moderate ratings, suggesting that metaphors in popular media are typically less elaborated or less consistent in resonance compared to more curated academic or clinical texts. A clear pattern emerges when examining smaller or more fragmented sources: for instance, the data by \citet{gibbs2002embodied} reveals extreme variability, with sentence-level ratings ranging from 0.00 to 6.75. 
This variability is amplified by the fact that many of these sources contribute only a handful of examples, making their average ratings less robust. Nonetheless, among sources with at least 40 annotated examples, the average ratings tend to cluster tightly between 1.99 and 2.41. This narrow band likely reflects the true central tendency for metaphor effectiveness in medical contexts. Ultimately, the observed distribution underscores how metaphor impact is context-sensitive: academic sources, clinical texts, and popular journalism differ in both intent and rhetorical strategy, while personal narratives, often emotionally charged, exhibit the highest degree of fluctuation.

We list below examples of highest rated and lowest rated metaphors:

\begin{description}
  \item[\textbf{Highest–rated}]%
    \begin{enumerate}[label=(\alph*)]
      \item \emph{It is inside the lungs that the virus turns nasty. It invades the millions of tiny air sacs in the lungs, causing them to become inflamed.}
      \item \emph{(about cell biology) Three-step theory of invasion.}
    \end{enumerate}

  \item[\textbf{Lowest–rated}]%
    \begin{enumerate}[label=(\alph*)]
      \item \emph{Two of its main activities—of the plasma membrane—are selective transport of molecules into and out of the cell.}
      \item \emph{(Of a person who has cancer) I have learned to let the little things go.}
    \end{enumerate}
\end{description}

%Themes for explo analysis:
%I can also check the themes now. We embed the metaphors and cluster them into themes according to semantic similarity

%Polarity:

%Are the metaphors positive or negative? The literatyre says that so etc

%Metaphoricity

%In general the average metaphoricity was and the most metaphorical ones were ...

%We also check the correlation between metaphoricity, polarity and themes. We find that ...

\section{Experimental setup}

As our primary contribution is the dataset itself rather than novel detection methods, we provide a baseline evaluation using state-of-the-art LLMs in zero-shot settings. This analysis establishes performance benchmarks for future method development while demonstrating the challenging nature of scientific metaphor detection. More sophisticated evaluation protocols (few-shot learning, fine-tuning, comparison with specialized metaphor detection models) represent important future work that our dataset enables (See Section 8.2).

\subsection{Evaluation metrics}

The evaluation process begins with establishing a standard from human annotations collected via Qualtrics surveys. As inter-annotator agreement is moderate, we can refer to a silver standard. For each metaphor detection item $q_i$, multiple human annotators provided binary judgments $R_i = \{r_1, r_2, \ldots, r_n\}$ where $r_j \in \{\text{yes}, \text{no}\}$. We compute vote counts as $\text{yes\_count}_i = \sum_{j=1}^{n} \mathbf{1}(r_j = \text{yes})$ and $\text{no\_count}_i = \sum_{j=1}^{n} \mathbf{1}(r_j = \text{no})$, where $\mathbf{1}(\cdot)$ is the indicator function. The majority label is determined as $\text{majority}_i = \text{yes}$ if $\text{yes\_count}_i > \text{no\_count}_i$, $\text{majority}_i = \text{no}$ if $\text{no\_count}_i > \text{yes\_count}_i$, and $\text{majority}_i = \text{tie}$ otherwise. Additionally, we calculate the confidence of each annotation as $\text{confidence}_i = \frac{\max(\text{yes\_count}_i, \text{no\_count}_i)}{\text{yes\_count}_i + \text{no\_count}_i}$, representing the proportion of annotators who agreed with the majority decision.
To account for varying levels of human agreement, we implement a confidence-based weighting scheme that assigns higher importance to items with stronger annotator consensus. The weight for each item is calculated as $w_i = 2 \cdot (\text{confidence}_i - 0.5)$ when $\text{confidence}_i > 0.5$, and $w_i = 0$ when $\text{confidence}_i = 0.5$ (ties). This linear mapping transforms confidence scores from the range $[0.5, 1.0]$ to weights in $[0.0, 1.0]$, ensuring that items with perfect consensus receive full weight while barely-majority cases receive minimal weight. Items where annotators were evenly split (ties) are effectively excluded from weighted calculations by receiving zero weight.
LLM predictions are evaluated against the human silver standard using both traditional and confidence-weighted metrics. Let $S = \{(y_i, \hat{y}_i, w_i) : \text{majority}_i \neq \text{tie}\}$ represent the set of non-tie predictions, where $y_i$ is the silver standard label, $\hat{y}_i$ is the model prediction, and $w_i$ is the confidence weight. Standard accuracy is computed as $\text{Accuracy} = \frac{1}{|S|} \sum_{i \in S} \mathbf{1}(y_i = \hat{y}_i)$, treating all items equally. The confidence-weighted accuracy is calculated as $\text{Weighted Accuracy} = \frac{\sum_{i \in S} w_i \cdot \mathbf{1}(y_i = \hat{y}_i)}{\sum_{i \in S} w_i}$, giving higher importance to items with stronger human consensus. Similarly, precision and recall metrics are computed both in standard form and with confidence weighting, where for class $c$, weighted precision is $\frac{\sum_{i \in S} w_i \cdot \mathbf{1}(y_i = c \land \hat{y}_i = c)}{\sum_{i \in S} w_i \cdot \mathbf{1}(\hat{y}_i = c)}$ and weighted recall is $\frac{\sum_{i \in S} w_i \cdot \mathbf{1}(y_i = c \land \hat{y}_i = c)}{\sum_{i \in S} w_i \cdot \mathbf{1}(y_i = c)}$.
Items where human annotators reached no consensus (ties) receive special treatment in our evaluation framework. During silver standard construction, tie items are identified and labeled but not assigned a definitive binary (yes/no) classification. In the weighting phase, these items receive zero weight ($w_i = 0$), effectively removing them from confidence-weighted calculations while preserving them in the dataset for transparency. During experimental model evaluation, tie items are completely excluded from all metric calculations, ensuring that models are only assessed on cases where human consensus exists. In our dataset, 134 items resulted in ties, leaving 589 items for evaluation. This exclusion strategy ensures that the evaluation focuses on cases with clear ground truth while avoiding penalizing models for predictions on inherently ambiguous examples where even human experts disagree.

%We use Accuracy and F1 score to evaluate the metaphor detection experiment.

%We evaluate metaphor detection as a binary classification task where models predict whether each scientific expression is metaphorical ("yes") or literal ("no"). Our gold standard is established through majority voting among human annotators collected via Qualtrics survey, where each item receives multiple human judgments. Items with tied human votes are accepted as valid annotations; in these cases, either answer is valid. We calculate standard binary classification metrics including precision, recall, and F1-score using scikit-learn's weighted averaging approach. The "Overall Accuracy" metric includes all items (treating human ties as automatically correct), while the primary evaluation metrics (Accuracy, Precision, Recall, F1) are computed only on items with clear human majority consensus. This approach ensures that our evaluation reflects genuine disagreement cases in the human annotation process while providing robust performance measures for model comparison. LLM responses undergo preprocessing to remove formatting artifacts (asterisks, underscores, backticks) and normalize case before comparison with human judgments.
%This methodology accounts for the inherent subjectivity in metaphor annotation while maintaining rigorous evaluation standards for automated detection systems.

\subsection{Models and parameters setup}

We used four LLMs exclusively through their APIs: GPT-4, o1-preview, o3-mini, Deepseek, Claude Opus 4. All experiments used default inference settings, with the sampling temperature fixed to~0 to obtain deterministic outputs. The sole exception is o1-preview, whose API mandates a default temperature of~1.

\subsection{Results}

Table~\ref{tab:metaphor_detection} presents standard evaluation metrics, while Table~\ref{tab:metaphor_detection_weighted} shows our confidence-weighted results. 

%o1-preview emerges as the strongest performer with 75.3\% confidence-weighted accuracy (vs 72.8\% standard), demonstrating reliable performance on cases with clear human consensus. Claude shows the largest improvement under confidence weighting (+2.9 percentage points), suggesting it excels at obvious metaphor detection while struggling more with borderline cases. Conversely, the unknown model shows minimal weighting benefit (+1.9\%), indicating frequent errors on clear-cut examples.

%All models exhibit conservative behavior with high precision for "yes" predictions (78-96\%) but lower recall (27-55\%), suggesting they reliably identify clear metaphors but miss many subtle instances. This pattern, combined with high "no" recall (90-98\%), indicates models err on the side of caution rather than over-detecting metaphors.

\begin{table}[htbp]
\centering
\caption{LLM Performance on scientific metaphor detection without weights. LLM Performance on scientific metaphor detection without weights. Precision, Recall and F1 are macro-averaged.}
\label{tab:metaphor_detection}
\begin{tabular}{lcccc}
\toprule
\textbf{Model} & \textbf{Acc} & \textbf{Prec} & \textbf{F1} & \textbf{Rec} \\
\midrule
o1-preview & 0.716 & 0.714 & 0.714 & 0.714 \\
Claude-Opus 4 & 0.711 & 0.746 & 0.707 & 0.725 \\
o3-mini & 0.706 & 0.785 & 0.695 & 0.727 \\
DeepSeek & 0.683 & 0.745 & 0.673 & 0.702 \\
GPT-4 & 0.655 & 0.785 & 0.695 & 0.727 \\
\bottomrule
\end{tabular}
\end{table}

\begin{table}[htbp]
\centering
\caption{LLM Performance on scientific metaphor detection with weighs.}
\label{tab:metaphor_detection_weighted}
\begin{tabular}{lcccc}
\toprule
\textbf{Model} & \textbf{wAcc} & \textbf{wPrec} & \textbf{wF1} & \textbf{wRec} \\
\midrule
o1-preview & 0.758 & 0.716 & 0.716 & 0.714 \\
Claude-Opus 4 & 0.755 & 0.756 & 0.705 & 0.721 \\
o3-mini & 0.752 & 0.799 & 0.690 & 0.706 \\
DeepSeek & 0.725 & 0.757 & 0.668 & 0.683 \\
GPT-4 & 0.690 & 0.776 & 0.626 & 0.655 \\
\bottomrule
\end{tabular}
\end{table}

\section{Discussion}

In this section, we discuss the results of the experimental setup and the potential of the dataset in computational metaphor research.

The relatively low inter-annotator agreement for binary rating reflects the inherent gradient nature of metaphoricity rather than annotation failure. This aligns with established findings in metaphor research: \citet{shutova2015design}, for instance, notes that moderate agreement is typical in metaphor annotation tasks due to the subjective nature of figurative language perception. Our Likert scale ratings (Pearson r = 0.441) capture indeed this gradient nature more effectively than binary judgments, suggesting that metaphoricity is better understood as a spectrum of literality rather than discrete categories.

Our exploratory analysis reveals a strong positive correlation between binary metaphoricity judgments and high metaphoricity ratings in scientific discourse. The substantial difference in literality ratings between metaphorical ($\mu$ = 3.41) and non-metaphorical expressions ($\mu$ = 0.16) suggests that annotators do perceive metaphors as having a low literality level. 

The disagreement patterns we identified also provide insights into the inherent challenges of metaphor annotation. Annotator judgments, in some cases, reveal genuine boundary cases involving scientific terminology with potential metaphorical readings, highly conventionalized metaphors, and domain-specific expressions where scientific expertise influences perception. These instances represent the most difficult cases for both human annotators and automated detection systems. Furthermore, such hard cases with low agreement tend to often represent the most theoretically interesting boundary phenomena rather than annotation failures.

%Table~\ref{tab:metaphor_detection} shows that for zero-shot scientific metaphor detection, o1-preview is the clear front-runner on every metric, achieving an overall score of~0.780 and an $F_{1}$ of~0.729.  Its advantage is explained less by a single standout dimension than by a balanced precision–recall profile: the model makes confident positive predictions while still recovering a substantial portion of true metaphors.
%By contrast, o3-mini and DeepSeek attain the highest precision figures but pay for this conservatism with markedly lower recall, yielding $F_{1}$ scores that trail the leader by roughly five percentage points.  Claude Opus 4 occupies an intermediate position, matching o1’s precision yet falling behind in recall and therefore in $F_{1}$. Finally, GPT-4 lags in both recall and overall accuracy, suggesting that instruction following alone is insufficient for the specialised task of scientific-metaphor detection.
The metaphoricity ranges in our dataset naturally enable confidence-weighted evaluation methodologies that account for varying levels of human consensus. By leveraging the degree of annotator agreement on each item, we can develop evaluation metrics that prioritize clear-cut cases while appropriately handling inherently ambiguous instances where human judgment varies.

Our confidence-weighted evaluation framework reveals that the consistent 3-4.6\% improvement across all models when weighted by human consensus indeed demonstrates that current LLMs perform systematically better on cases where humans strongly agree, while struggling disproportionately with ambiguous instances.

This pattern has important implications for practical applications: o3-mini's largest weighting benefit (+4.6\%) suggests it could serve reliably in high-confidence scenarios while requiring additional safeguards for borderline cases. o1-preview's balanced performance across both weighted and standard metrics indicates more robust handling of metaphor ambiguity, making it suitable for applications requiring consistent performance across diverse linguistic contexts. We attribute these models’ success to the fact that they are tuned for deliberate reasoning in few-token budgets; their internal chain-of-thought appears particularly effective for short, domain-specific classification zero-shot prompts like ours.

The conservative precision-recall profiles observed across all models (high precision but low recall for metaphor detection) reflect a systematic bias toward literal interpretation. This case suggests that LLMs adopt a cautious decision boundary, labelling a sentence as metaphorical only when strongly lexical cues (e.g.\ \textit{``battle,'' ``storm,''} or explicit anthropomorphism) are present.
While this reduces false positives, it may limit utility in applications requiring comprehensive metaphor identification, such as literary analysis or patient communication assessment.

%The relatively low inter-annotator agreement (Fleiss' κ = 0.25) reflects metaphoricity's gradient nature rather than annotation failure. Our Likert scale ratings (Pearson r = 0.441) capture this gradient more effectively than binary judgments, supporting theoretical arguments that metaphoricity exists on a continuum rather than as discrete categories.

%Boundary cases involving scientific terminology with potential metaphorical readings, highly conventionalized metaphors, and domain-specific expressions represent the most challenging instances for both human annotators and automated systems. These cases, while theoretically interesting, highlight the need for domain expertise in annotation and potentially specialized training approaches for computational systems.

 %This makes them reliable but blind to subtler metaphors grounded in scientific language.  Claude displays the opposite tendency: it is more willing to call borderline cases metaphors, improving recall but suffering false positives whenever technical terms are figurative only in specialised subfields.

Therefore, the MCC dataset surfaces cases that even frontier LLMs find non-trivial, making it a valuable stress-test for future metaphor-aware language technology.

\subsection{Applications and Future Directions}
The proposed MCC dataset opens several promising avenues for practical applications and research. In computational linguistics, the annotated metaphors can improve metaphor detection, understanding, and generation systems by providing training data that captures both metaphorical status and its range, alongside source and target domains. The dataset's potential extends to personalized communication tools, such as in education, but also particularly in medical settings where controlled metaphor selection could enhance patient understanding and engagement. Promising avenues include (i) fine-tuning or continued pre-training on the MCC dataset; and (ii) integrating symbolic ontologies with LLMs to bias inference toward structured, yet context-based metaphor understanding and analysis.

For scientific writing tools and educational applications, the dataset could support the development of writing assistants that suggest appropriate metaphors for complex scientific concepts.

Future work could also expand the dataset to track the consequences of specific metaphorical mappings, enabling controlled studies of metaphor effectiveness in scientific communication. This could lead to the creation of dynamic, evidence-based metaphor repositories that inform real-time writing assistance tools. Additionally, investigating how metaphor perception varies across different scientific domains and expertise levels could further refine our understanding of figurative language in specialized discourse.

\subsection{Limitations and future work}
Our dataset is limited to English-language scientific texts, which restricts the generalizability of findings to other languages where metaphorical expressions and their metaphoricity perception may differ significantly. Additionally, while our dataset provides a substantial foundation with scientific metaphors and metaphoricity ratings, expanding the corpus with more metaphorical expressions and more fine-grained annotation dimensions (e.g. quality ones: clarity, creativity, appropriateness) would enhance its utility for diverse research applications.
The relatively low inter-annotator agreement, while not uncommon in metaphor annotation tasks, presents challenges for establishing reliable silver standards in the field of scientific metaphors. 
Furthermore, the dataset represents a snapshot of contemporary scientific writing and may not capture evolving metaphorical conventions or cultural variations in metaphor perception. Longitudinal studies tracking metaphor usage and quality perception over time could reveal important trends in scientific communication practices.

\section{Data availability}
The MCC dataset and the user-annotated data is publicly available on GitHub at \url{https://anonymous.4open.science/r/medical-metaphors-corpus-86B7/README.md}. A permanent Zenodo DOI will be provided upon paper acceptance to comply with anonymity requirements.

\section{Ethics statement}
All data was collected from publicly available sources with no private medical information accessed. Human annotation involved 40 voluntary participants who provided informed consent and could withdraw at any time. The dataset contains no personally identifiable information and represents published discourse. We acknowledge limitations including English-language and Western cultural bias, and commit to responsible data sharing practices. All data was collected in accordance with fair use and fair dealing provisions for academic research. Academic sources are used under scholarly fair use exemptions for criticism, analysis, and research purposes. News media excerpts fall within UK fair dealing provisions for research and quotation. Social media content was previously collected by researchers following appropriate ethical guidelines for publicly available discourse. The dataset uses only short excerpts and sentence-level examples rather than substantial portions of original works, supporting fair use claims under the transformative purpose and limited quantity factors.

\section{Conclusion}
In this work, we have introduced the \textbf{Medical Metaphors Corpus} (\textsc{MCC}), the first openly released resource that captures metaphorical language across the breadth of medical and biological discourse. Spanning 792 sentences and 82 distinct metaphor types, each enriched with human-curated binary metaphoricity labels, graded (0–7) metaphoricity scores, and curated source–target mappings, \textsc{MCC} fills a critical gap between general‐domain metaphor datasets and the needs for new use cases for NLP. Using \textsc{MCC} as a benchmark, we evaluated five LLMs under zero‐shot conditions.  Our evaluation using confidence-weighted metrics demonstrates that while o1-preview achieved the strongest performance, all models show systematic weaknesses in handling metaphorical ambiguity. In fact, the consistent improvement under confidence weighting reveals that current LLMs perform reliably on clear-cut cases but struggle disproportionately with borderline instances.

%These findings highlight both the promise and limitations of current LLM approaches to metaphor detection. While models like o1-preview show encouraging baseline performance, the gap between weighted and standard metrics indicates substantial room for improvement in handling metaphorical ambiguity.

%Although \texttt{o1-preview} achieved the strongest overall performance, every model struggled with subtle, domain‐specific metaphors: precisely the expressions that clinicians, researchers, and patients encounter daily. These findings show cutting‐edge language models remain brittle when literal and figurative meanings intertwine in technical prose. 
Thus, \textsc{MCC} provides a new testbed for LLMs, which still struggle in metaphor processing tasks.  

Looking ahead, we envision expanding \textsc{MCC} both horizontally, to other scientific metaphors, domains and languages, and \emph{vertically}, by adding richer annotation aspects such as emotional valence, explanatory clarity, and multimodality to power controllable metaphor generation, for example in clinical settings.

%%%%%%%

%\iftaclpubformat
%Note that you will need to provide both a single-spaced and a double-spaced
%version; see \S \ref{ssec:layout}.

%If you promised to provide code or data at submission, specific instructions for
%how to access such resources must be provided.  (Typically, a URL to a stable,
%resource-specific site suffices.)

%All URLs should be manually checked to verify that they
%lead to a valid webpage, and to the site that was intended.
%\fi

%\iftaclpubformat

\bibliography{tacl2021}
\bibliographystyle{acl_natbib}

%\iftaclpubformat

\onecolumn

\appendix

%\fi

\end{document}